\documentclass[conference]{IEEEtran}
\IEEEoverridecommandlockouts
\usepackage{cite}
\usepackage{amsmath,amssymb,amsfonts}
\usepackage{algorithmic}
\usepackage{graphicx}
\usepackage{url}
\usepackage{textcomp}
\usepackage{xcolor}
\usepackage{subcaption}

\begin{document}

\title{Stable Diffusion with Continuous-time Neural Networks}

\author{\IEEEauthorblockN{Andras Horvath}
\IEEEauthorblockA{Faculty of information Technology and Bionics\\
\textit{Peter Pazmany Catholic University}\\
Budapest, Hungary\\
horvath.andras@itk.ppke.hu}
}
\maketitle

\begin{abstract}

Stable diffusion models have ushered in a new era of advancements in image generation, currently reigning as the state-of-the-art approach, exhibiting unparalleled performance. The process of diffusion, accompanied by denoising through iterative convolutional or transformer network steps, stands at the core of their implementation. Neural networks operating in continuous time naturally embrace the concept of diffusion, this way they could enable more accurate and  energy efficient implementation.

Within the confines of this paper, my focus delves into an exploration and demonstration of the potential of celllular neural networks in image generation. I will demonstrate their superiority in performance, showcasing their adeptness in producing higher quality images and achieving quicker training times in comparison to their discrete-time counterparts on the commonly cited MNIST dataset.

\end{abstract}


\section{Introduction}
Image generation has witnessed remarkable progress in recent years, fueled by the advancement of deep learning techniques. Among the numerous approaches, Stable Diffusion Algorithms \cite{rombach2022high} have emerged as a promising paradigm for generating high-quality and coherent images. Unlike traditional generative models, these algorithms harness the power of diffusion processes to produce images gradually, offering inherent stability and control over the generation process.

As a latent diffusion model, Stable Diffusion belongs to the category of deep generative artificial neural networks. Its primary function revolves around generating intricate images based on random or conditioned input such as text descriptions, but it extends its utility to other tasks like inpainting, outpainting and producing image-to-image translations with guidance from textual prompts\cite{tumanyan2023plug}.

The fundamental concept underpinning stable diffusion revolves around the progressive degradation of images into Gaussian noise, achieved through incremental noise addition. This process retains its reversibility, and by approximating its inverse, structured images can be synthesized from initially random noise inputs.

In theoretical terms, this procedure embodies image diffusion. However, in practical implementations, given the prevalence of discrete-time architectures, this is translated into a sequence of steps predominantly facilitated by convolutional networks.

In the context of this progression, the number of iterations and the algorithm's time-step consistently stand as adjustable parameters. This is owing to the need to approximate continuous-time diffusion using convolution operations.

Fortunately, the realm of continuous-time neural networks, such as cellular neural networks\cite{chua1993cnn} can present a solution in the form of cellular neural networks. These networks operate in an uninterrupted temporal framework, with their feedback template inherently embodying the essence of diffusion.

My objective is to ascertain and demonstrate that the integration of continuous-time diffusion into actual diffusion models can yield models of greater accuracy while maintaining equivalent complexity, in contrast to their discrete-time counterparts. It's worth noting that diffusion, in conjunction with analog hardware, could potentially offer even greater efficiency. However, I have to acknowledge that this aspect has not been explored within the scope of our paper.

\let\thefootnote\relax\footnotetext{This work was supported by the European Union’s Horizon EU research and innovation program, Horizon EU PHASTRAC (https://phastrac.eu) project under Grant no. 101092096.}

\section{Stable diffusion}

Probabilistic models known as stable diffusion models are specifically designed to acquire knowledge about a data distribution $p(x)$ through a progressive denoising process of a normally distributed variable. This process essentially involves learning the inverse of a fixed Markov Chain with a predefined length $T$, where for a specific timestep $t=1 \dots T$:

\begin{equation}
  L_{DM} = \mathbb{E}_{E(X), \epsilon } ~ N(0,1), t \left [   \left \|  q - q _o (x_t,t) \right \|_2^2 \right ]
\end{equation}

\begin{figure}[htp]
\centering
\includegraphics[width=3.5in]{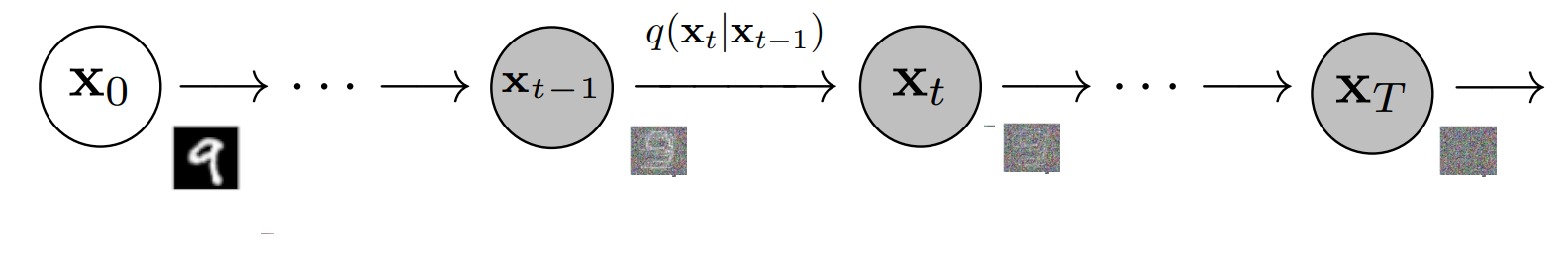}
\caption{Depiction of the diffusion which generates random Gaussian noise from an input image. The approximated inverse of this process can be used for image generation which was demonstrated in \cite{rombach2022high}.}
\label{fig:diffproc}
\end{figure}

These models can be seen as a set of denoising autoencoders, denoted as $E(xt, t)$, where $t$ ranges from 1 to T. Each autoencoder is assigned equal importance and is trained to predict a denoised version of its input $x_t$. Here, $x_t$ represents a noisy variant of the original input $X$ and $q$ notes a distance metric between two states of the process. A simplified version of this Markovian process is depicted in Figure \ref{fig:diffproc}.

\begin{figure}[htp]
\centering
\includegraphics[width=3.5in]{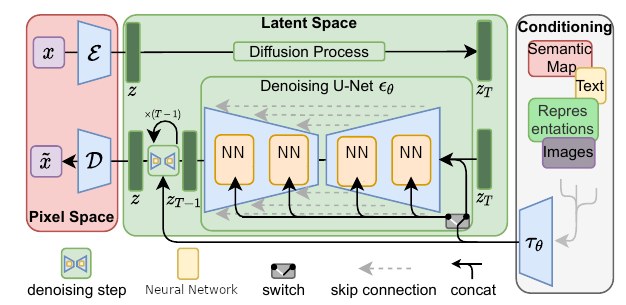}
\caption{Typical architecture of a Latent Diffusion Model where the inverse of the diffusion step - which is also a diffusion process- is approximated by a series of discrete operations (self-attentions or convolutions). The original image was taken from \cite{rombach2022high}.}
\label{fig:diffarch}
\end{figure}

LDMs (Latent Diffusion Models) offer a versatile and computationally efficient approach for synthesizing diverse image modalities based on diffusion. Intriguingly, it was also observed that LDMs trained in vector quantization regularized latent spaces can achieve high sample quality, despite the slightly diminished reconstruction capabilities of vector quantization regularized first-stage models compared to their continuous counterparts \cite{zhang2023regularized}. A typically applied stable diffusion architecture is depicted in Figure \ref{fig:diffarch}.

As it can be seen in this figure the inverse of the diffusion process is approximated by a series of discrete operations, meanwhile it is continuous in nature and in this paper I will demonstrate that can be better approximated with a continuous time network which implements real diffusion.

\section{Cellular neural networks and Memristive Cellular neural networks}

Over the past decades, CellNNs have demonstrated their utility in tasks involving image and pattern generation \cite{arena2000collective}. In these applications, fixed parameters were the norm, with no parameter training. Furthermore, these applications typically utilized single-channel networks. Nevertheless, the realm of CellNNs showcased their potential to seamlessly generate complex patterns through proper parametrization.

CellNNs also stand out as highly promising architectures for the energy-efficient and timely execution of intricate data processing tasks. Such computing platforms are currently in high demand, addressing the rigorous demands of Internet-of-Things (IoT) and Edge Computing applications \cite{horvath2017cellular}.

In two noteworthy works by Fulop and colleagues \cite{fulop2020template} and \cite{fulop2021application}, the authors underscored the effectiveness of gradient-based approaches not only in training convolutional networks but also in optimizing the programming templates of CellNNs using these popular methods.

Meanwhile convolutional networks are predominantly implemented on Boolean, discrete-time devices, CellNNs carry out all computations in continuous time and in an analog fashion.

The state variable of the CellNN model can be defined by the following differential equation:
{\small
\begin{equation}\label{CellNNEquation}
\begin{array}{c}
  \dfrac{dx_{i,j}(t)}{dt}= 
- x_{i,j}(t)+\sum\limits_{k,l \in S_{i,j}} A_{i,j,k,l} y[x_{i,j,k,l}(t)] + \\
+ \sum\limits_{k,l \in S_{i,j}} B_{i,j,k,l} u_{i,j,k,l}(t) +Z_{i,j}
\end{array}
\end{equation}
}

Here, $x$ denotes the state variable, $u$ represents the input, and $A$, $B$, $Z$ are the connecting weights (templates of the array). Furthermore, $i$, $j$, $k$, and $l$ act as index variables that select a cell within a neighborhood radius ($S$) of the processing array.

The output function $y$ of each cell is determined by the following equation:
\begin{equation}
    y = f_{out}(x)=\frac{1}{2} \left | x+1 \right | -  \frac{1}{2} \left | x-1 \right |
\end{equation}

However, this non-linearity fails to provide gradients for training when the output values fall below -1 or exceed 1. To circumvent this limitation, the authors introduced the Leaky CellNN Non-linearity, defined as follows:
\begin{equation}\label{EQleaky}
y=
\left\{\begin{matrix}
-1 + \alpha(x+1) & ,if  & -1 > x \\ 
x & ,if&  -1 \leq x \leq 1 \\ 
1 + \alpha(x-1) & ,if &  1<x
\end{matrix}\right.
\end{equation}
Here, $\alpha$ typically assumes a low value, such as $0.01$ in their and also my experiments.

To implement a convolutional layer in CellNNs, one can discretize Equation \ref{CellNNEquation} and set all values in matrix $A$ to zero. 

%Another distinct operator in ConvNNs is the pooling operation, which selects the maximum intensity in a small neighborhood of the image. This operation downscales the image, introduces additional non-linearity to the network, and is computationally straightforward. However, it does not find a counterpart in CellNNs, where the spatial dimensions of the input remain consistent in each layer.

It becomes evident that by employing an $A$ template with all elements of matrix $A$ set to zero, a singular convolutional effect akin to that of a single convolutional layer is achieved. Simultaneously, incorporating the $A$ template significantly expands the range of feasible functions, empowering the network to leverage these capabilities for implementing intricate diffusion processes which can be directly beneficial in case of approximating diffusion processes.

This practical manifestation of the diffusion phenomenon is supported by evidence that CellNNs can effectively mimic genuine diffusion processes, such as heat diffusion \cite{hills2014cellular} or energy propagation  \cite{hadad2007application}.

\subsection{Memristive Cellular neural networks}

\begin{figure}
\centering
\subfloat{\includegraphics[width=1.0\linewidth,height=.4\linewidth]{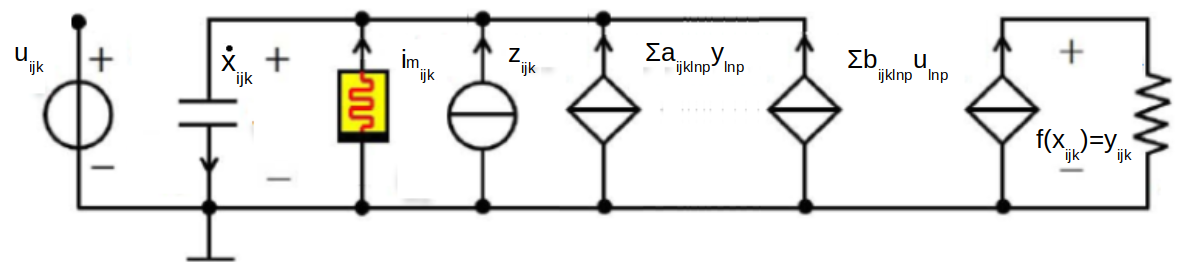}}
\caption{Illustration of the circuit of a multi-channel M-CellNN cell $C_{i,j,k}$ The
input (output) stage lies on the leftmost (rightmost) side of the circuit.
}\label{FigCells}
\end{figure}

The construction of each layer within the proposed Deep Neural Networks (DNNs) is based on a multi-layer M-CellNN. In this configuration, each processing unit (illustrated in Fig. \ref{FigCells} circuit-theoretic representation) possesses two degrees of freedom encoded into capacitor voltage and memristor state, respectively.

For the second-order cell of a multi-channel M-CellNN, the ordinary differential equation (ODE) governing the time evolution of the voltage across its capacitance includes an additional term, specifically the current through the memristor $i_{m_{i,j,k}},$ in comparison to the state equation (\ref{CellNNEquation}) of the first-order cell $C_{i, j, k}$ in a traditional multi-channel Cellular Neural Network (CellNN). This differential equation is as follows:

{\small
\begin{equation}\label{MemristiveCellNNEquation}
\begin{array}{c}
  \dot{x}_{i,j,k}= 
- x_{i,j,k}\\
+\sum\limits_{l,n,p \in S_{i,j,k}} a_{i,j,k,l,n,p} y_{l,n,p} + \\
+ \sum\limits_{l,n,p \in S_{i,j},k} b_{i,j,k,l,n,p} u_{l,n,p} +z_{i,j,k}-  i_{m_{i,j,k}},
\end{array}
\end{equation}
}
where $i_{m}$denotes the current flowing through $m_{i,j,k}$. The
dynamics of this memristor are expressed through the DAE set:
\begin{equation}
\dot{m_{i,j,k}} = g(m_{i,j,k}, v_{m_{i,j,k}})
\end{equation}

\begin{equation}
i_{m_{i,j,k}}= G(m_{i,j,k},v_{m_{i,j,k}}) \cdot x_j
\end{equation}

Here, $i_{m}$ denotes the current flowing through $m_{i,j,k}$. The dynamics of this memristor are expressed through the Differential-Algebraic Equation (DAE) set:

\begin{equation}
\dot{m_{i,j,k}} = g(m_{i,j,k}, v_{m_{i,j,k}})
\end{equation}

\begin{equation}
i_{m_{i,j,k}}= G(m_{i,j,k},v_{m_{i,j,k}}) \cdot x_j
\end{equation}

Where $v_{m_{i,j,k}} = x_{i,j,k}$ and $v_{m_{i,j,k}}$ respectively represent the voltage and state of the memristor $m_{i,j,k}$ in cell $C_{i,j,k}$. The specific form of the state evolution and memductance functions was provided in Section III for each of the two memristor models studied in this research.

The initial conditions for the two states of the cell $C_{i,j,k}$, namely the capacitor voltage $x_{i,j,k}$ and memristor state $m_{i,j,k}$, are denoted as $x_{i,j,k}(0)$ and $m_{i,j,k}(0)$, respectively. The functions $g(m_{i,j,k}, v_{m_{i,j,k}})$ and $G(m_{i,j,k}, v_{m_{i,j,k}})$ are defined according to a specific memristor model.

M-CellNN cell features only one additional element, namely a memristor, in comparison to a classical CellNN cell. The inclusion of a memristor per cell does not significantly increase power consumption or the manufacturing complexity of the hardware implementation of the array. However, as demonstrated in the remainder of the paper, enriching the dynamics of each cellular unit by introducing a memristor device in parallel with the respective capacitor leads to an enhanced computing engine. This engine, based on multiple layers of multi-channel M-CellNNs, outperforms traditional implementations of multi-layer Deep Neural Networks (DNNs) with similar complexity in both classification and segmentation problems.

In my experiments Tantalum oxide (TaO\textsubscript{x}) memristors were investigated, which are a highly favored choice in the realm of resistance random access memory (ReRAM) devices for non-volatile memory applications, largely due to their promising physical and electrical properties. An experimentally-derived compact model has been introduced for nanoscale TaO\textsubscript{x} devices in \cite{strachan2013state}, capturing their intricate slow/fast switching dynamics, which depend on initial conditions and input settings.

This model is an extended memristor model that is voltage-controlled, described by a first-order nonlinear ordinary differential equation and a nonlinear Ohm's Law that depends on both state and input variables. In the equations:
\begin{subequations}
\begin{eqnarray}
\begin{array}{l}
\frac{dm}{dt} =  g(m, v_m) = A \cdot  sinh(\frac{v_m}{\sigma_{off}}) \cdot  e^{\frac{-m^2_{off}}{m2}} e^{\frac{1}{1+\beta i_m v_m}} \cdot  \\ 
\textrm{step}(-v_m)+ B \cdot  sinh(\frac{v_m}{\sigma_{on}}) \cdot  e^{\frac{-m^2}{m^2_{on}}} e^{\frac{i_m v_m}{\sigma_p }} \cdot \textrm{step}(v_m)
\end{array}
\\
\begin{array}{l}
i_m = v_m G(m,v_m),
\end{array}
\end{eqnarray}\label{TAOeq}
\end{subequations}

$m$ represents the dimensionless state variable, with values normalized and confined to the closed interval [0, 1].
$v_m$ and $i_m$ denote the memristor voltage and current, respectively.
$\sigma_{on}$ and $\sigma_{off}$ are parameters with physical units of voltage.
$step(\cdot)$ is the step function.
$G(m,v_m)$ is the memconductance function.
A comprehensive explanation of the significance of each parameter in equations (10a) and (10b), along with the values employed in the simulations,  can be found in Table I in \cite{demirkol2021compact}.

\section{Measurements and Results}

I have selected a Latent Diffusion Model as a baseline model, as shown in Figure 2, to serve as my reference implementation. The PyTorch implementation of this architecture is available on GitHub at the following link: \url{https://github.com/TeaPearce/Conditional_Diffusion_MNIST}.

The structure of this architecture closely resembles that of a UNET, with two downscaling and two upscaling blocks. Each block incorporates a pair of residual blocks, collectively hosting 128 features.

The initial training of this network was carried out using both the MNIST dataset, consisting of 28x28 resolution images representing the ten digits and the CIFAR-10 dataset with resolution 32x32 containing 10 classes of three channeled colored images.

The training process spanned ten epochs, with batch sizes of 16 instances. This implementation serves as a foundational baseline reference, essential for comparison and evaluation purposes.

For comparison, I used an identical architectural layout, with the only difference being the replacement of conventional residual blocks with CellNN and M-CellNN layers. This substitution ensures equivalence in terms of complexity and the number of operations performed. It is important to note that the residual blocks involve discrete-time operations, while my adaptation incorporates continuous-time diffusion.

Due to the constraints of contemporary digital hardware in efficiently executing continuous-time operations, I resorted to simulating this process using the TorchDiffEq Python module. The numerical approximation of the differential equations involved 100 steps, with a time step of 0.01 for each iteration.

It is crucial to acknowledge that this approximation significantly increases the computational complexity of the algorithm. However, it is worth noting that specific hardware configurations are capable of efficiently executing cellular neural networks involving voltage diffusion. In such scenarios, the computation aligns seamlessly with the underlying physics and the primary objective of this paper was not to assess the efficiency of a physical device but to compare the two implementations in a simulated environment.

For the sake of completeness and
reproducibility, a comprehensive description of each of the three architectures, under
the zooming lens in this study, including details on its fixed parameter setting and on the
outcome of one of its trainings, may be found in the Github repository:
-Link was removed for the sake of anonymity-.

\begin{figure}[htp]
  \centering
    \begin{subfigure}[b]{.13\textwidth}
        \centering
        \includegraphics[width=\linewidth]{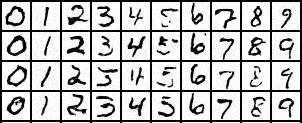}
    \end{subfigure}%
    \hspace{0.5pt}
      \centering
    \begin{subfigure}[b]{.13\textwidth}
        \centering
        \includegraphics[width=\linewidth]{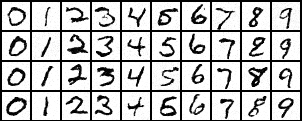}
    \end{subfigure}%
    \hspace{0.5pt}
      \centering
    \begin{subfigure}[b]{.13\textwidth}
        \centering
        \includegraphics[width=\linewidth]{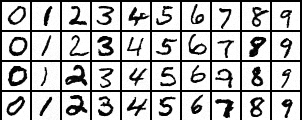}
    \end{subfigure}%
    \\
    \vspace{0.5pt}
     \begin{subfigure}[b]{.13\textwidth}
        \centering
        \includegraphics[width=\linewidth]{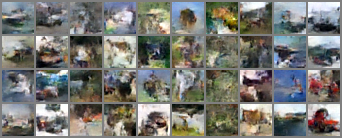}
    \end{subfigure}%
    \hspace{0.5pt}
      \centering
    \begin{subfigure}[b]{.13\textwidth}
        \centering
        \includegraphics[width=\linewidth]{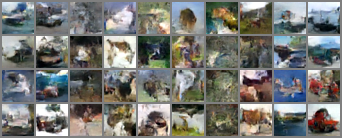}
    \end{subfigure}%
    \hspace{0.5pt}
      \centering
    \begin{subfigure}[b]{.13\textwidth}
        \centering
        \includegraphics[width=\linewidth]{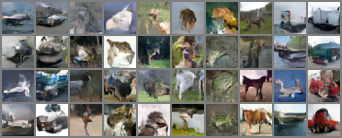}
    \end{subfigure}%
\caption{These images offer qualitative outcomes, presenting samples for comparative assessment. The images in the top row are results from models trained on the MNIST, and in the bottom row on the CIFAR-10 datasets. Images in the left column showcase samples produced by conventional stable diffusion models incorporating convolutional blocks. In the middle samples generated with CellNN and in the right column samples generated by the TaO\textsubscript{x} M-CellNN are displayed. }
\label{fig:qulaitative}
\end{figure}

A comparison between the three implementations reveals noticeable qualitative distinctions, as visualized in Figure \ref{fig:qulaitative}.

It is important to note that while convolutional networks can achieve improved results through the use of more complex architectures and longer training iterations, the objective here was to compare two architectures under similar conditions and complexities.

As illustrated in Figure \ref{fig:qulaitative}, it is clear that the cellular neural network has delivered superior sample quality. Furthermore, the memristive network performs even better, producing MNIST samples that are more regular and CIFAR samples that are more detailed and sharp, with clearly defined object borders, as well as cleaner backgrounds.

Quantitatively comparing generative models presents challenges due to the inherent nature of generative modeling, which involves randomness in generated samples. Therefore, direct comparisons at a sample level are complex. The focus often shifts toward analyzing diverse samples rather than directly comparing specific instances.

Conventional metrics, such as pixel-based distances, prove inadequate in this scenario as they assess discrepancies between identical objects. For instance, evaluating two images of the same digit, even if they differ significantly in pixel composition, does not necessarily imply that one of the samples exhibits poor quality.

To address this, I selected the Fréchet Inception Distance (FID) score \cite{vahdat2021score}, a commonly used metric to quantitatively evaluate and compare the quality of generated images against a reference dataset. The FID score measures the perceptual similarity between generated images and real images by utilizing features extracted from a pretrained deep neural network model, such as Inception-v3 or similar architectures. In my measurements, I used a pretrained LeNet-5 network trained on the MNIST dataset and achieving $98\%$ accuracy and ResNet-50 arhcitecture was used on CIFAR-10 with $81\%$ accuracy.
 
In essence, the FID score considers both the mean and covariance of feature distributions. It evaluates the quality of individual generated samples and the diversity across the set of generated images. Lower FID scores indicate higher quality and similarity to the real dataset.
The FID scores on the MNIST and CIFAR dataset using convolutional, CellNN and M-CellNN architectures can be seen in Table \ref{tab:Values}.
Similarly to the qualitative results, the measurements affirm that memristive networks can yield the lowest FID scores, indicating the highest image quality within this setup.

\begin{center}
\begin{table}
\centering
\begin{tabular}{ccc}
\hline
  \begin{tabular}[c]{@{}l@{}} Method\end{tabular} &
  \begin{tabular}[c]{@{}l@{}} MNIST FID $\downarrow$\end{tabular}  &
  \begin{tabular}[c]{@{}l@{}} CIFAR FID $\downarrow$\end{tabular} 
  \\
  \hline
\begin{tabular}[c]{@{}c@{}}
CONV\end{tabular} &
16.2  & 2.78\\ 
\begin{tabular}[c]{@{}c@{}}
CellNN\end{tabular} &
 13.4  & 2.45\\ 
 \begin{tabular}[c]{@{}c@{}}
M-CellNN\end{tabular} &
12.8 & 2.27\\ 
  \hline
\end{tabular}
\caption{This table provides evaluations for the three investigated architectures: convolutional (CONV), cellular (CellNN), and memristive cellular neural network (M-CellNN). The two columns present FID scores obtained from the MNIST and CIFAR-10 datasets. The performance of these methods is assessed using FID scores, where lower values signify superior performance. Each number is the average of ten independent trainings.}\label{tab:Values}
\end{table}
\end{center}

\section{Conclusion}

This paper effectively demonstrates the suitability of CellNNs and memristive grids in latent diffusion models. The utilization of CellNNs has been proven to offer substantial benefits in terms of sample quality. Although these improvements may be characterized as modest, they are undeniably tangible. Notably, this enhancement can be attributed to the departure from the traditional paradigm, where diffusion is approximated as a sequence of convolutions. Instead, the approach taken in this study leverages the direct execution of diffusion processes. The employment of memristive elements can further increase the image generation quality of CellNNs.

The findings presented here emphasize the potential of CellNNs and M-CellNNs as valuable tools in the realm of latent diffusion modeling. The ability to move away from convolutional approximations and embrace the authenticity of diffusion processes could open up new avenues for further innovation and optimization in this field. As the field of neural networks continues to advance, the insights gained from this study could contribute to the ongoing development of techniques that bridge the gap between continuous-time diffusion and its discrete-time approximations.

\section*{Acknowledgement}\label{ACK}
Tis project has received funding from the EU’s Horizon program under Project No. 
101092096 (PHASTRAC) which is gratefully acknowledged.
This research has also been partially supported by the Hungarian Government by the following grants: 2018-1.2.1-NKP00008: Exploring the Mathematical Foundations of Artificial Intelligence and TKP2021\_02-NVA-27 – Thematic Excellence Program.

\bibliographystyle{ieeetr}
\bibliography{cnn}

\end{document}